\documentclass[conference]{IEEEtran}
\IEEEoverridecommandlockouts
\usepackage{cite}
\usepackage{amsmath,amssymb,amsfonts}
\usepackage{algorithmic}
\usepackage{graphicx}
\usepackage{textcomp}
\usepackage{xcolor}

\usepackage{subfigure}
\usepackage{hyperref}
\usepackage{doi}

\usepackage{flushend}

\usepackage{fancyhdr}
\fancyfootoffset{1cm}
\fancypagestyle{firstpage}{
\lfoot{\footnotesize \vspace{-1cm} This work has been published: D. Huczala, T. Kot, J. Mlotek, J. Suder and M. Pfurner, "An Automated Conversion Between Selected Robot Kinematic Representations," \emph{2022 10th International Conference on Control, Mechatronics and Automation (ICCMA),} Belval, Luxembourg, 2022, pp. 47-52, doi: \href{https://doi.org/10.1109/ICCMA56665.2022.10011595}{10.1109/ICCMA56665.2022.10011595}.
\\ \vspace{0.2cm} \copyright2022 IEEE. Personal use of this material is permitted. Permission from IEEE must be obtained for all other uses, in any current or future media, including reprinting/republishing this material for advertising or promotional purposes, creating new collective works, for resale or redistribution to servers or lists, or reuse of any copyrighted component of this work in other works.
} 
}

\def\BibTeX{{\rm B\kern-.05em{\sc i\kern-.025em b}\kern-.08em
    T\kern-.1667em\lower.7ex\hbox{E}\kern-.125emX}}
\begin{document}

\title{An Automated Conversion Between Selected Robot Kinematic Representations\\
\thanks{This article has been elaborated under support of the project Research Centre of Advanced Mechatronic Systems, reg. no. CZ.02.1.01/0.0/0.0/16\textunderscore019/0000867 in the frame of the Operational Program Research, Development and Education. This article has been also supported by specific research project SP2022/67 and financed by the state budget of the Czech Republic.}
}

\author{

\IEEEauthorblockN{Daniel Huczala}
\IEEEauthorblockA{\textit{Department of Robotics} \\
\textit{VSB-Technical University of Ostrava}\\
Ostrava, Czech Republic \\
daniel.huczala@vsb.cz \href{https://orcid.org/0000-0002-7398-7825}{\includegraphics[scale=0.1]{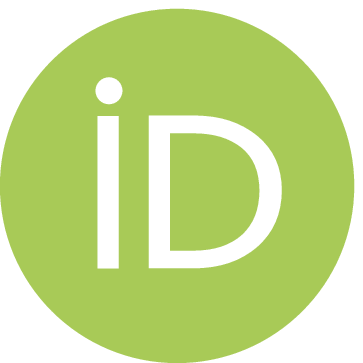}}}
\and
\IEEEauthorblockN{Tomáš Kot}
\IEEEauthorblockA{\textit{Department of Robotics} \\
\textit{VSB-Technical University of Ostrava}\\
Ostrava, Czech Republic \\
\href{https://orcid.org/0000-0003-2357-806X}{\includegraphics[scale=0.1]{orcid.eps}}}
\and
\IEEEauthorblockN{Jakub Mlotek}
\IEEEauthorblockA{\textit{Department of Robotics} \\
\textit{VSB-Technical University of Ostrava}\\
Ostrava, Czech Republic \\
\href{https://orcid.org/0000-0003-2070-5908}{\includegraphics[scale=0.1]{orcid.eps}}}

\and 

\hspace{3cm}
\IEEEauthorblockN{Jiří Suder}
\IEEEauthorblockA{
\hspace{3cm} \textit{Department of Robotics} \\
\hspace{3cm}
\textit{VSB-Technical University of Ostrava}\\
\hspace{3cm}
Ostrava, Czech Republic \\
\hspace{3cm}
\href{https://orcid.org/0000-0002-4692-1329}{\includegraphics[scale=0.1]{orcid.eps}}}

\and

\IEEEauthorblockN{Martin Pfurner}
\IEEEauthorblockA{\textit{Unit of Geometry and Surveying} \\
\textit{University of Innsbruck}\\
Innsbruck, Austria \\
\href{https://orcid.org/0000-0003-1988-2202}{\includegraphics[scale=0.1]{orcid.eps}}}

}

\maketitle

\begin{abstract}
This paper presents a methodology that forms an automated tool for robot kinematic representation conversion, called the RobKin Interpreter. It is a set of analytical algorithms that apply basic linear algebra tools that can analyze an input serial robot representation, express the joints globally in matrix form, and map to other representations such as standard Denavit-Hartenberg parameters, Roll-Pitch-Yaw angles with translational displacement, and Product of Exponentials with a possibility to generate a URDF (Universal Robot Description Format) file from any of them. It works for revolute and prismatic joints and can interpret even arbitrary kinematic structures that do not have orthogonally placed joints and often appear in flexible robotic systems. The aim of the proposed methods is to facilitate the necessary switches between representations due to software compatibility, calculation simplification, or cooperation.
\end{abstract}


\section{Introduction}\label{sec:introduction}

\thispagestyle{firstpage}

There are a few types of representation to describe robotic structures and related parameters to provide versatility in terms of kinematic and dynamic calculations. Their objective is to specify the position and orientation of the robot end-effector or joints, which generally gives six degrees of freedom in the $\mathbb{R}^3$ space for every joint. In case of the most simple and most commonly used joints -- prismatic and revolute, there is one variable specifying the joint displacement or rotation and affecting the position of the subsequent joints and the end-effector. Simply, the joints can be represented as a right-hand coordinate frame with $x, y, z$ position vector and $a, b, c$ orientation vector. The orientation vector can be specified by the Tait-Bryan angles or better known as Roll-Pitch-Yaw (RPY) angles \cite{Cho2021} around the $x$, $y$ and $z$ axes, respectively. These angles may also be described as an alternative to Euler angles \cite{Pio1966, Cashbaugh2018}. Note that there is a difference between the RPY and Euler angles values because of the reference frame used. However, the principle is the same and both representations have a major lack in terms of mathematical singularities \cite{Ang1987}. The combination of six parameters (RPY rotations and XYZ translations around/along given axes) will be addressed as RPY-XYZ representation in this study. 

Recently, the Robot Operating System (ROS) has increased the usage of the Universal Robot Description Format (URDF), which is an XML (eXtensible Markup Language) file format to describe the kinematics and dynamics of robots in a tree structure. The URDF is based on RPY-XYZ representation, and, besides ROS, it is implemented in other software such as Gazebo, CoppeliaSim, and MATLAB.

Due to the singularity issues mentioned above, some representations prefer the matrix form over the RPY-XYZ six-vector. They use the matrix representation of the Special Euclidean group \textit{SE}(3) that is suitable for representing the rigid body orientation in space. Probably the most common representation of robot kinematics is the Denavit--Hartenberg convention \cite{Denavit1955} using four parameters to describe the displacement between two coordinate frames (joints) that are placed in a special order as the convention prescribes. Another widely known representation is based on Screw Theory \cite{Ball1876}, where the joint axes are represented as normalized twists defined by a directional vector and a linear velocity vector at the origin \cite{lynch2017modern}. It is also called the Product of Exponentials (PoE) representation, and it is an extension to the Plücker coordinates \cite{Joswig2013}. 

Usually, one uses a representation that, for some reason or benefit, prefers over the others. However, sometimes it is necessary to switch between representations for multiple reasons, i.e., algorithm compatibility, software support, cooperation with other team members, simplification, and calculation efficiency. At this point, such a change can become an issue if one has never experienced another representation, and a lot of errors and long time-related debugging may follow. This problem even extends to the case of manipulators with arbitrary or flexible kinematic structures, which often appear as output after their synthesis \cite{Husty2007, Hauenstein2017} for specific tasks. The joint axes of these manipulators can be freely placed, creating curved links \cite{Du2015} or even complete curved \cite{brandstotter2015curved} and flexible manipulators \cite{Clark2020, Mlotek2022}. In addition, for such structures, it becomes challenging to determine any representation, let alone to switch between. This was the crucial motivation for us to create an automated converter between the common robot representations for serial manipulators.

There are a few papers dealing with the automatic determination of a robot representation. Corke \cite{Corke2007} presents an algorithm for obtaining DH parameters from simple rotations and translations. Kang et al. \cite{Kang2019} uses the DH parameters as input to obtain a corresponding URDF file. However, it works only with revolute joints. In the appendix of the book by Lynch and Park \cite{lynch2017modern} an algorithm can be found to derive PoE elements -- screws and transformation matrix between the base frame and the tool frame representing a manipulator -- defined by DH parameters. Wu et al. \cite{Wu2017, Wu2019n} introduced a method to obtain DH parameters from PoE for revolute, prismatic, and helical joints. By default, their algorithm assumes to use the tool frame placed according to DH convention and eventually to add a transform to it. This makes it difficult to switch between the representations back and forward.

This paper presents analytical algorithms which are able to map between \emph{standard} DH parameters, PoE, and RPY-XYZ representations with the possibility of generating a simple version of a URDF file out of them. The core of the presented approach is to express any joint $i$ of a robot in the home configuration as a homogeneous transformation matrix $\mathbf{J_i} \in$ \textit{SE}(3) in the base frame of the robot. We call it Global Joint Description and it is referred to as GJD later on. From this description of the joints it is possible to obtain any of the three chosen joint representations. Unlike the existing methods mentioned in the previous paragraph, the main advantage of the GJD approach is its possible extension to other representations (e.g., the modified DH convention by Craig \cite{craig2009introduction}), since it is based on the matrix form of the \textit{SE}(3) group rather than by direct calculations between two selected representations. This brings a versatile tool that is freely available on an open source repository \cite{Huczala2021git}, with future plans for its extensions. It supports serial manipulators with revolute and prismatic joints from 1 to $n$ number of joints. Section \ref{sec:methodology} presents the analytical methods that provide switches from different representations to and from GJD and in Section \ref{sec:examples} there are two examples of robotic manipulators which are mapped in other representations.

\section{Methodology}\label{sec:methodology}
The key idea in all algorithms presented is to geometrically analyze the representation of the input robot of a serial manipulator and find such transformation matrices whose coordinates correspond to the joints of the input kinematics in the home configuration. The algorithms implemented in MATLAB are available online \cite{Huczala2021git} and to run them one needs to use Corke's Spatial Math Toolbox \cite{Corke2021smt, Corke17n} additionally. The visualization figures, which can be found later in this paper, were obtained using MATLAB Robotics Systems Toolbox \cite{matlab-rst}.

In this study, we denote the transformation matrix between two frames $\mathbf{T} \in$ \textit{SE}(3) 
\begin{equation}
\small
    \label{eq:T}
    \mathbf{T} = \\
    \begin{bmatrix}
         &  &  &  \\ 
        \vec{n} & \vec{o} & \vec{a} & \vec{t} \\
         &  &  & \\
        0 & 0 & 0 & 1\\ 
    \end{bmatrix}
\end{equation}
where~$\vec{n}$ (normal) is the X axis vector, the~$\vec{o}$ (orientation) is Y axis vector, the~$\vec{a}$ (approach) is Z axis vector, and~the $\vec{t}$ (translation) is the coordinate vector position between the origins of the two frames.

In the following, we will have a look at the three different representations and their relation to a common matrix description. For a single transformation, the RPY-XYZ description are six parameters, three of them for the translation and the other three for the orientation. In case of DH-parameters, the position of one frame with respect to another can be expressed using four parameters only. Both of these representations can be expressed in matrix form, which will be shown later, and the forward kinematics is calculated as a sequence of partial transformations between consecutive axes. On the contrary, a screw representing a joint is a $6\times1$ vector denoted as
\begin{equation}
    \mathcal{S} = 
    \begin{bmatrix}
        \vec{\omega}\\ 
        \vec{v}  \\
    \end{bmatrix} =
    \begin{bmatrix}
        \vec{\omega}\\ 
        -\vec{\omega} \times q  \\
    \end{bmatrix}
\end{equation}
in the case of a revolute joint, $\vec{\omega}$ is $3\times1$ unit vector representing the direction of its axis and $\vec{v}$ is $3\times1$ vector representing the linear velocity of the screw axis in the base frame. It can be calculated as $-\vec{\omega} \times q$, where $q$ is any point on the axis. For prismatic joints $\vec{\omega} = [0, 0, 0]^T$ and $\vec{v}$ is the unit vector representing the direction of the linear displacement. To calculate the forward kinematics, along with the screws, the PoE representation requires also to define a matrix $\mathbf{M}$, which is a transform from the base of a robot to its end-effector in the home configuration of the manipulator~\cite{lynch2017modern}.

\subsection{DH parameters to GJD}

The input is a table of standard DH parameters. Depending on the task, a tool frame transform $\mathbf{T}_t$ shall be added if the DH convention cannot express the full transformation between the last joint and the tool frame on its own. 

At first, the partial transformation matrices $\mathbf{A}_{i-1,i}$ are obtained following the DH convention.

\begin{equation}
\small
    \label{eq:A_i2}
    \mathbf{A}_{i-1,i} = \mathbf{R}_{z_{i-1}}(\theta_{i})\mathbf{T}_{z_{i-1}}(d_{i})\mathbf{T}_{x_i}(a_i)\mathbf{R}_{x_i}(\alpha_i)
\end{equation}

where $\mathbf{R}$ is a pure rotation, $\mathbf{T}$ is a pure translation, and the corresponding indexes denote the axes of the transformation.
%$\mathbf{R}_z(\theta_i)$ is the matrix of pure rotation with angle $\theta_i$ around $z_{i-1}$ axis, %$\mathbf{T}_z(d_i)$ is the matrix of pure translation with displacement $d_i$ along $z_{i-1}$ axis, %$\mathbf{T}_x(a_i)$ is the matrix of pure translation with displacement $a_i$ along $x_i$ axis, and  %$\mathbf{R}_x(\alpha_i)$ is the matrix of pure rotation with angle $\alpha_i$ around $x_i$ axis. 
The transformation matrix $\mathbf{J}_i$ of the $i_{th}$ joint represented in the base frame (GJD) is obtained by $\prod_{j=1}^{i}\mathbf{A}_{j-1,j}$, that is, the chain of previous partial transforms.

%\begin{equation}
%\small
%    \label{eq:J_i}
%    \mathbf{J}_{i} = \mathbf{A}_{01}\mathbf{A}_{12}...\mathbf{A}_{i-1,i}
%\end{equation}

For $n$ joints there are $\mathbf{J}_{1..n}$ transforms and frame $\mathbf{J}_{n+1}$ representing the end-effector. %and may be containing the additional transformation $\mathbf{T}_t$. If the DH convention cannot fully express the end-effector frame, the differing matrix $\mathbf{T}_t$ is added.

\begin{equation}
\small
    \label{eq:J_n1}
    \mathbf{J}_{n+1} = \mathbf{J}_{n} \mathbf{T}_t
\end{equation}
%If the convention can express the end-effector frame, $\mathbf{T}_t$ becomes an identity.

\subsection{PoE to GJD}

Obtaining the GJD of the PoE representation is not straightforward. The inputs for this algorithm are the matrix $\mathbf{M}$ and the joint screws $\mathcal{S}_{1..n}$. For later simplification, we determine such GJD transformation matrices so they also fulfill the DH convention, therefore coordinates frames %$\mathbf{J}_{1..n}$ transforms 
representing the joints are placed along common perpendiculars (the nearest distance) between consecutive joint axes.

A line can be represented by a vector and a point. The directional vector is known from the screw $\mathcal{S}$ as $\vec{\omega}$ (revolute) or $\vec{v}$ (prismatic). Therefore, we can define a line $l_i$ corresponding to the joint axis of the joint $i_{th}$ as

\begin{equation}
    l_i:
    \begin{cases}
        (\vec{\omega}_i, q_i),  & \text{if joint } i \text{ is revolute}  \\
        (\vec{v}_i, q_i),       & \text{if joint } i \text{ is prismatic}
    \end{cases}
\end{equation}

The point $q_i$ is any point on the joint axis. If the joint $i$ is revolute, the closest point on the screw axis to the origin can be expressed as $q_i = \vec{\omega_i} \times \vec{v_i}$. If the joint is prismatic, its screw $\mathcal{S}_i$ has zero moment on the origin, therefore, we place the point in the position of the previous joint $q_i = q_{i-1}$. This is one of the major limitations of the PoE representation when converting to another one, and shall be taken in mind if using prismatic joints. However, the forward kinematic problem yields the same solution for the structures nevertheless where the prismatic joint is placed. 

To achieve matrices $\mathbf{J_i}$ in order to describe the joint axes equal to once in the DH convention we have to distinguish between 4 cases. The joint axes $(i-1)$ and $i$ are:

\begin{enumerate}
    \item Skew or parallel lines -- the $i_{th}$ joint is placed on the intersection of the common normal and $l_i$. Its $x$ axis is placed along the normal pointing from line $l_{i-1}$ to $l_i$. In the case of parallel lines, the common normal is determined from the origin of $(i-1)_{th}$ joint.
    \item Coincident lines -- the frames are coincident $\mathbf{J}_{i} = \mathbf{J}_{i-1}$.
    \item Coincident lines but $\vec{\omega}_{i-1}$ has opposite orientation than $\vec{\omega}_{i}$ -- the frame is rotated around $x$ axis $\mathbf{J}_{i} = \mathbf{J}_{i-1}\mathbf{R}_x(\pi)$.
    \item Intersecting with an angle -- $x$ axis equals the normalized cross product of the previous $z$ axis ($\vec{a}_{i-1}$) and the current $z$ axis ($\vec{\omega_i}$): $\vec{a}_{i-1} \times \vec{\omega_i}$. The frame is placed on the intersection point of $l_{i-1}$ and $l_i$
\end{enumerate}

The output transformation matrix $\mathbf{J}_{n+1}$ between the tool and base frames is equal to $\mathbf{M}$.

\subsection{RPY-XYZ to GJD and GJD to RPY-XYZ}

RPY-XYZ uses the partial transformation matrices between frames, while with GJD we represent the same coordinate frames in relation to the base of a robot. To switch between partial and global transformations, standard calculation methods \cite{Ang1987} were applied along with the functions of Spatial Math Toolbox. It is based on the multiplication of the rotational matrices for each axis (RPY angles) with added XYZ displacement in the total homogeneous transformation.

\subsection{GJD to DH parameters}

This approach to obtain DH parameters between two lines is described in \cite{Huczala2021} and is applied between all partial transformations. If there is a tool that does not obey the DH convention (automatically tested by the algorithm), it is obtained as follows. 

Obtain a frame $\mathbf{J}'_{n+1}$ that tries to reach the input tool frame $\mathbf{J}_{n+1}$ with DH parameters. Pre-multiply it by $\mathbf{J}_n$ to provide a transformation to the base frame.

\begin{equation}
\small
    \mathbf{J}'_{n+1} = \mathbf{J}_{n}\mathbf{R}_{z_n}(\theta_{n+1})\mathbf{T}_{z_n}(d_{n+1})\mathbf{T}_{x_{n+1}}(a_{n+1})\mathbf{R}_{x_{n+1}}(\alpha_{n+1})
\end{equation}

Calculate the partial tool frame transform $\mathbf{T}_t$.

\begin{equation}
\small
    \mathbf{T}_t  = \mathbf{J}'^{-1}_{n+1} \mathbf{J}_{n+1} 
\end{equation}

The table of DH parameters and $\mathbf{T}_t$ are the output.

\subsection{GJD to PoE}

The transformation $\mathbf{M}$ between the end-effector and the base frame of a robot is equal to the globally expressed (GJD) tool frame $\mathbf{J}_{n+1}$. The screw vectors $\mathcal{S}$ are obtained in the case of prismatic joints as

\begin{equation}
    \label{eq:S_ip}
    \mathcal{S}_{i} = 
    \begin{bmatrix}
        0 \\
        0 \\
        0 \\
        \vec{a}_{i} \\
    \end{bmatrix} \\
\end{equation}

where $\vec{a}_{i}$ is the $z$ vector coordinate of the frame $\mathbf{J_i}$. For revolute joints, the screw vector reads

\begin{equation}
    \label{eq:S_ir}
    \mathcal{S}_{i} = 
    \begin{bmatrix}
       \vec{a}_{i} \\
       -\vec{a}_{i} \times \vec{t}_{i} \\
    \end{bmatrix}, 
\end{equation}

where $\vec{t}_{i}$ is any point on the screw axis, in this case the origin of the $\mathbf{J_i}$ frame.

\subsection{Export to URDF}

The Universal Robot Description Format is based on the RPY-XYZ representation as default. Our Matlab script can generate a raw URDF file based on the input RPY-XYZ table, where the first row stands for the base transform of the robot and the last row stands for the end-effector transform. Rows 2 to $n+1$ represent the partial transforms between $n$ number of joints. The dynamic parameters of the robot are not included, but the output XML file can be freely adjusted afterwards.

\section{Examples and Results}\label{sec:examples}

To demonstrate and evaluate the algorithms, we chose two kinematic structures. The standard (meaning orthogonal or parallel joint axes) RRPR structure has orthogonal and parallel links, the other 3R structure has joint axes placed arbitrary. Both structures have displaced end-effector tool. In the used acronyms, R stands for revolute and P for prismatic joints of the serial manipulator.

\begin{figure*}
    \centering
    \subfigure[Input DH parameters]
    {
        \includegraphics[width=0.31\textwidth]{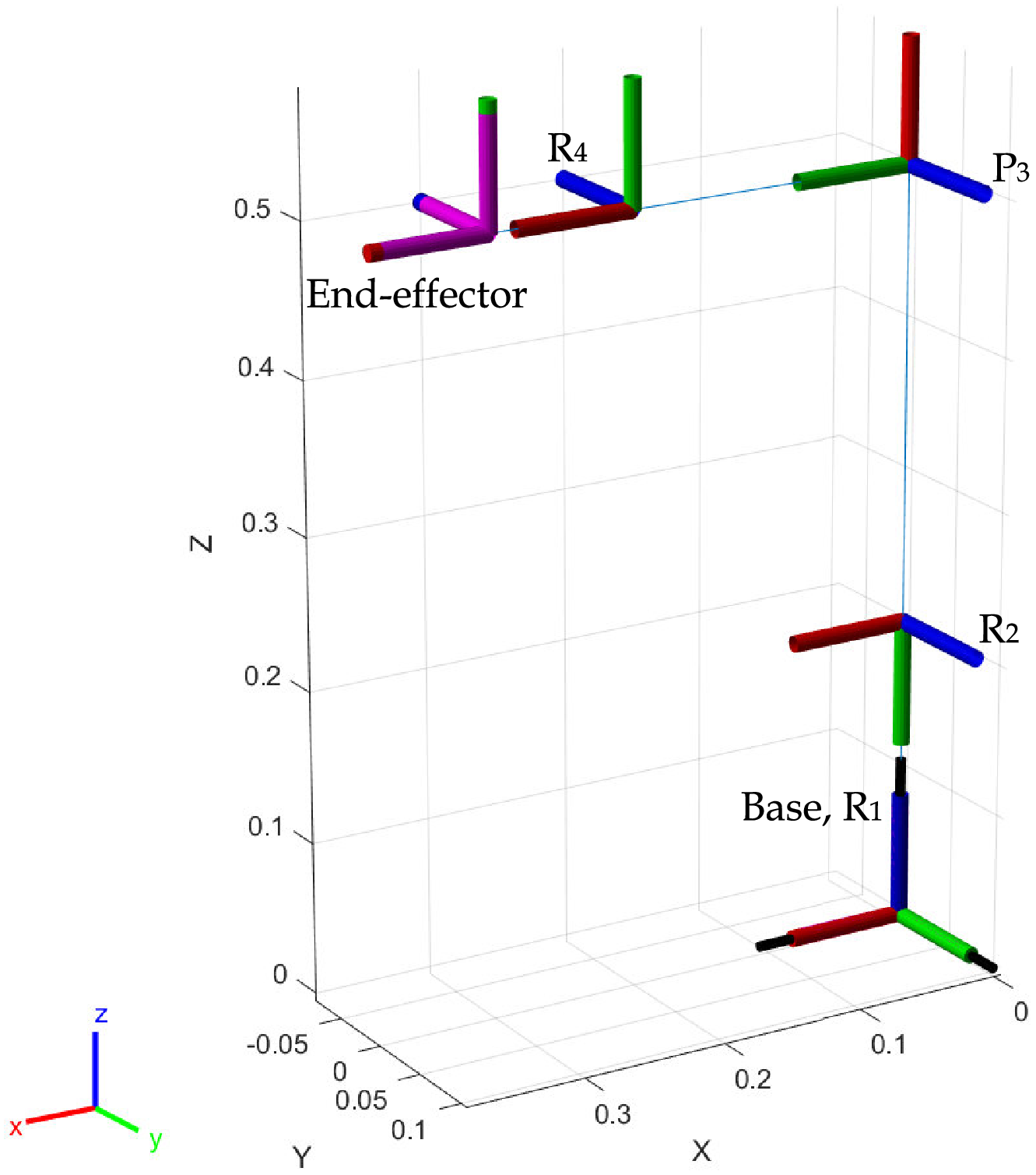}
    }
    \subfigure[Output RPY-XYZ visualization of screws (PoE)]
    {
        \includegraphics[width=0.31\textwidth]{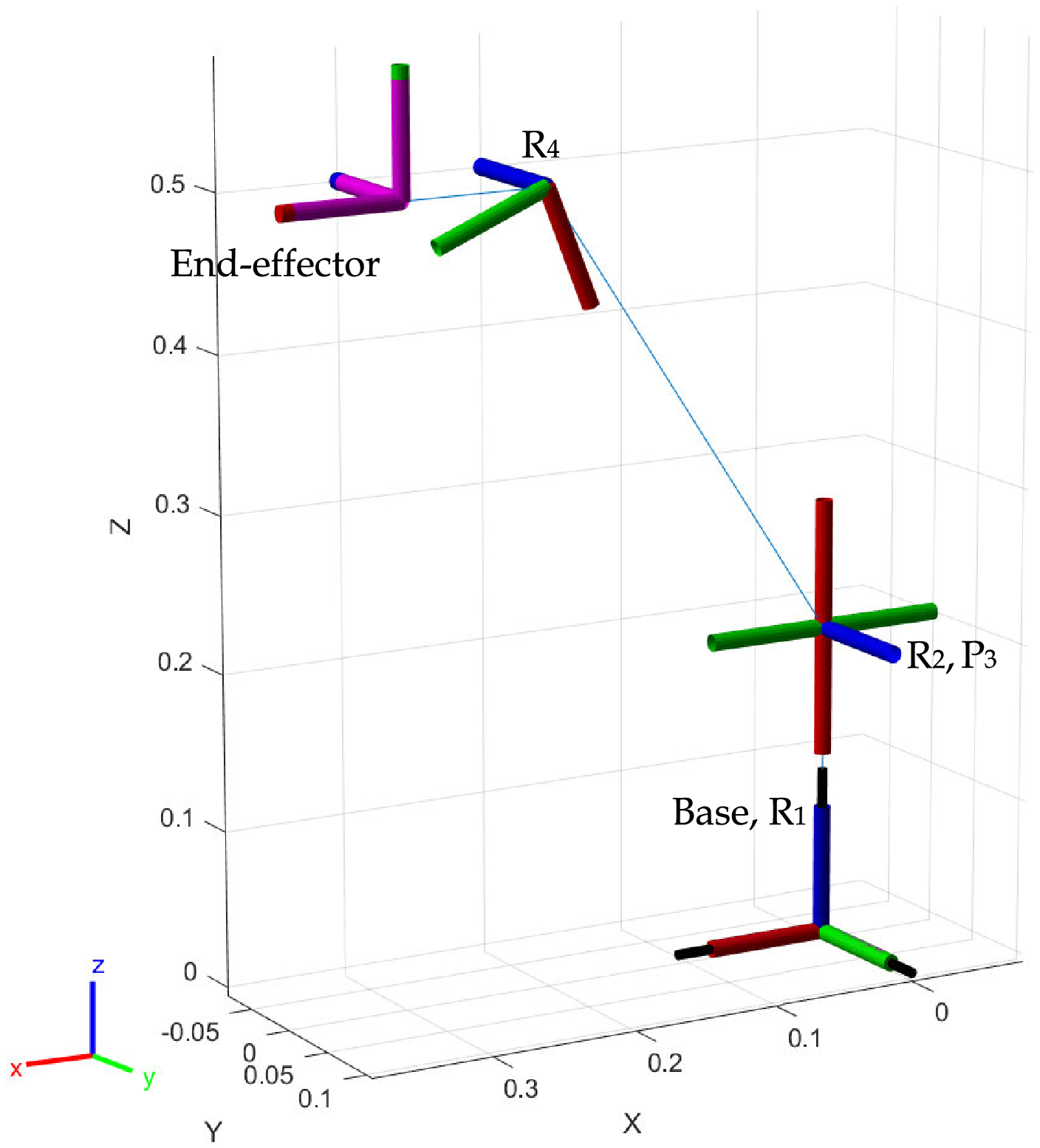}
    }
    \subfigure[Output DH parameters]
    {
        \includegraphics[width=0.31\textwidth]{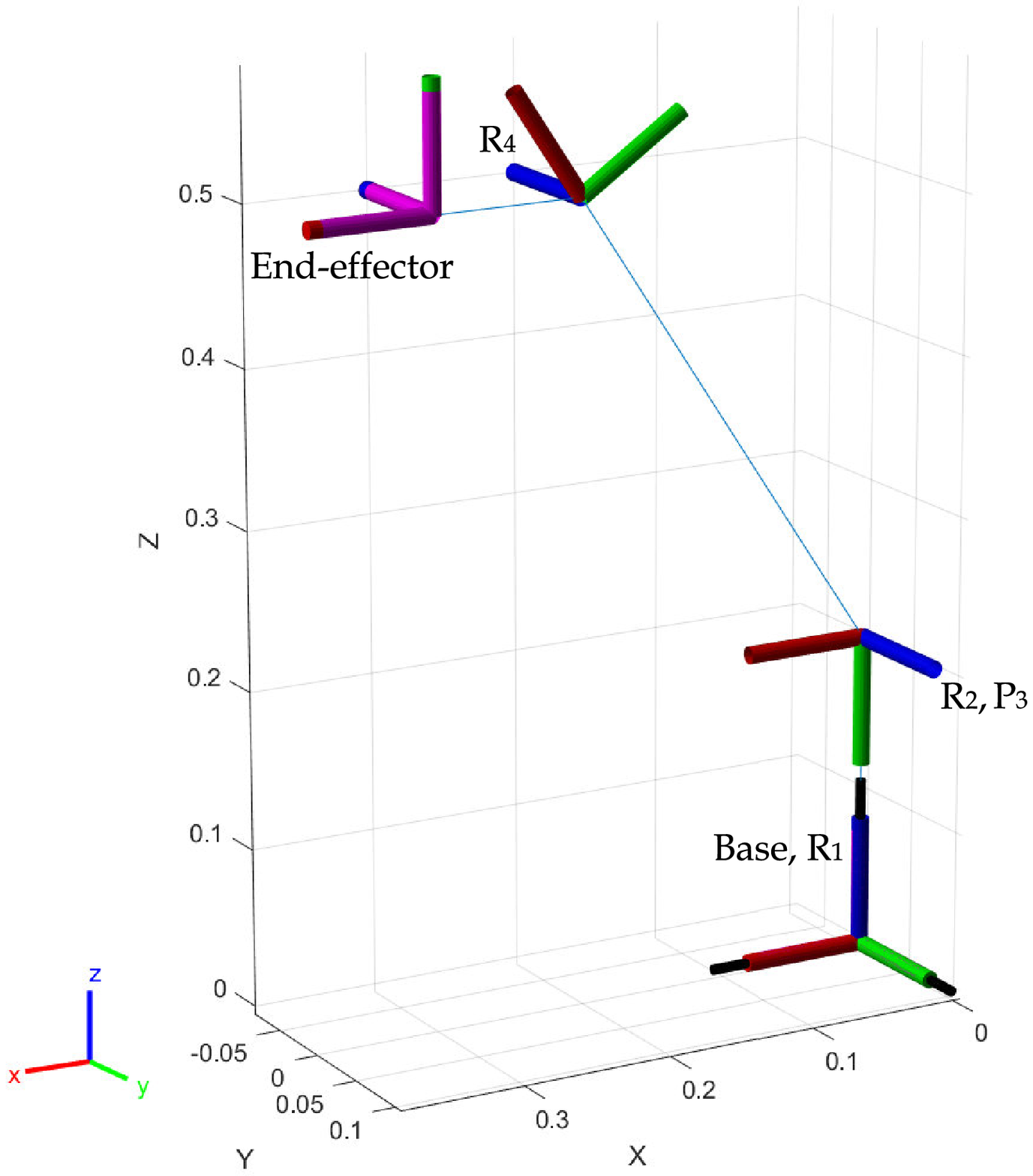}
    }
    \caption{The visualization of the RRPR robot at home configuration with different representations; joint axes have blue color; tool frame has pink color.}
     \label{fig:results_rrpr}
\end{figure*}

\subsection{Standard RRPR structure}

The input representation for this RRPR robot is shown in Table \ref{tab:dh_RRPR}, they are standard DH parameters with no additional tool offset.

\begin{table}[h]
%\footnotesize
\caption{The DH parameters of the RRPR robot; default offset of joint variables is in brackets}
\vspace{-8pt}
\label{tab:dh_RRPR}
\begin{center}
\begin{tabular}{c | c c c c }
 i    &  $a_i$ [m] &  $d_i$ [m] &  $\alpha_i$ [rad] &  $\theta_i$ [rad] \\
 \hline\hline
 0 & 0 & 0 & 0 & 0 \\ 
 \hline
 1 & 0 & 0.2 & --$\pi$/4 & $q_1$(0) \\
 \hline
 2 & 0.3 & 0 & 0 &  $q_2$(--$\pi$/4)\\
 \hline
 3 & 0.2 &  $q_2$(0) & $\pi$ & $\pi$/4\\
 \hline
 4 & 0.1 & 0 & 0 &  $q_4$(0) \\ 
\end{tabular}
\end{center}
\end{table}

The methods of Section \ref{sec:methodology} were applied to obtain RPY-XYZ, the result is presented in Table \ref{tab:rpy_RRPR}. 

\begin{table}[h]
%\footnotesize
\caption{The RPY-XYZ of the RRPR robot}
\vspace{-8pt}
\label{tab:rpy_RRPR}
\begin{center}
\begin{tabular}{c c c c c c}
 R [rad] &  P [rad] &  Y [rad] &  X [m] & Y [m] &  Z [m] \\
 \hline\hline
 0 & 0 & 0 & 0 & 0 & 0 \\ 
 \hline
 0 & 0 & 0 & 0 & 0 & 0 \\ 
 \hline
 --$\pi$/2 & 0 & 0 & 0 & 0 & 0.2\\
 \hline
 0 & 0 & --$\pi$/2 & 0 & --0.3 & 0 \\
 \hline
 $\pi$ & 0 & $\pi$/2 & 0 & 0.2 & 0\\
 \hline
 0 & 0 & 0 & 0.1 & 0 & 0 \\ 
\end{tabular}
\end{center}
\end{table}

The PoE representation of the robot given in Table \ref{tab:dh_RRPR} was determined with these results: 
\begin{equation}
\small
\label{eq:m_rrpr}
    \mathbf{M} = \\
    \begin{bmatrix}
        1 & 0 & 0 & 0.3 \\ 
        0 & 0 & -1 & 0 \\
        0 & 1 & 0 & 0.5 \\
        0 & 0 & 0 & 1\\ 
    \end{bmatrix}
\end{equation}

\begin{equation}
\small
\label{eq:screws_rrpr}
    \mathcal{S}_1 = 
    \begin{bmatrix}
        0 \\ 
        0 \\
        1 \\ 
        0 \\
        0 \\ 
        0 \\
    \end{bmatrix}
    \quad
    \mathcal{S}_2 = 
    \begin{bmatrix}
        0 \\ 
        1 \\
        0 \\ 
        -0.2 \\
        0 \\ 
        0 \\
    \end{bmatrix}
    \quad
    \mathcal{S}_3 = 
    \begin{bmatrix}
        0 \\ 
        0 \\
        0 \\ 
        0 \\
        1 \\ 
        0 \\
    \end{bmatrix}
    \quad
    \mathcal{S}_4 = 
    \begin{bmatrix}
        0 \\ 
        -1 \\
        0 \\ 
        0.5 \\
        0 \\ 
        -0.2 \\
    \end{bmatrix}
\end{equation}

\begin{figure*}
    \centering
    \subfigure[Input DH parameters]
    {
        \includegraphics[width=0.31\textwidth]{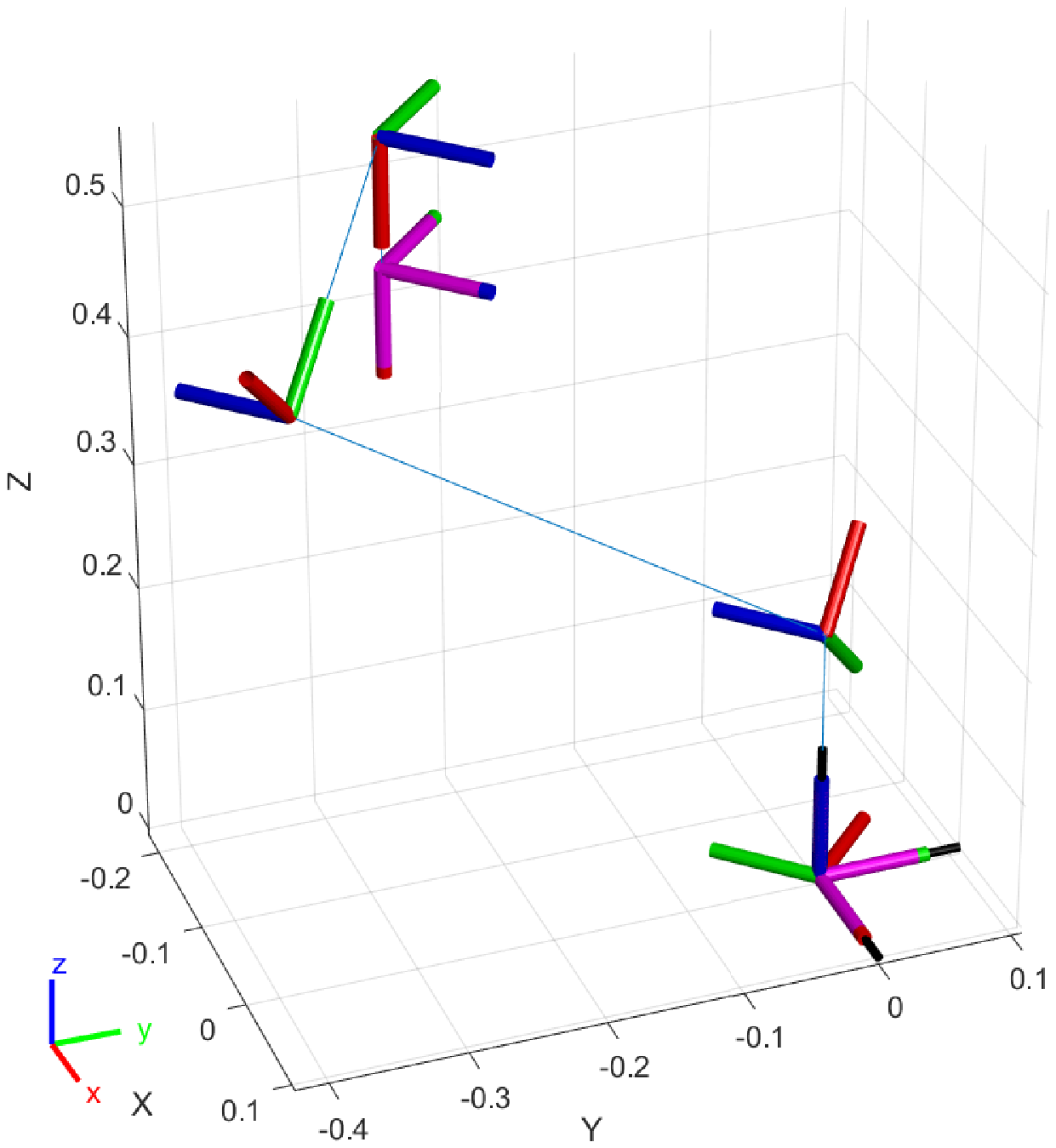}
    }
    \subfigure[Output RPY-XYZ visualization of PoE]
    {
        \includegraphics[width=0.32\textwidth]{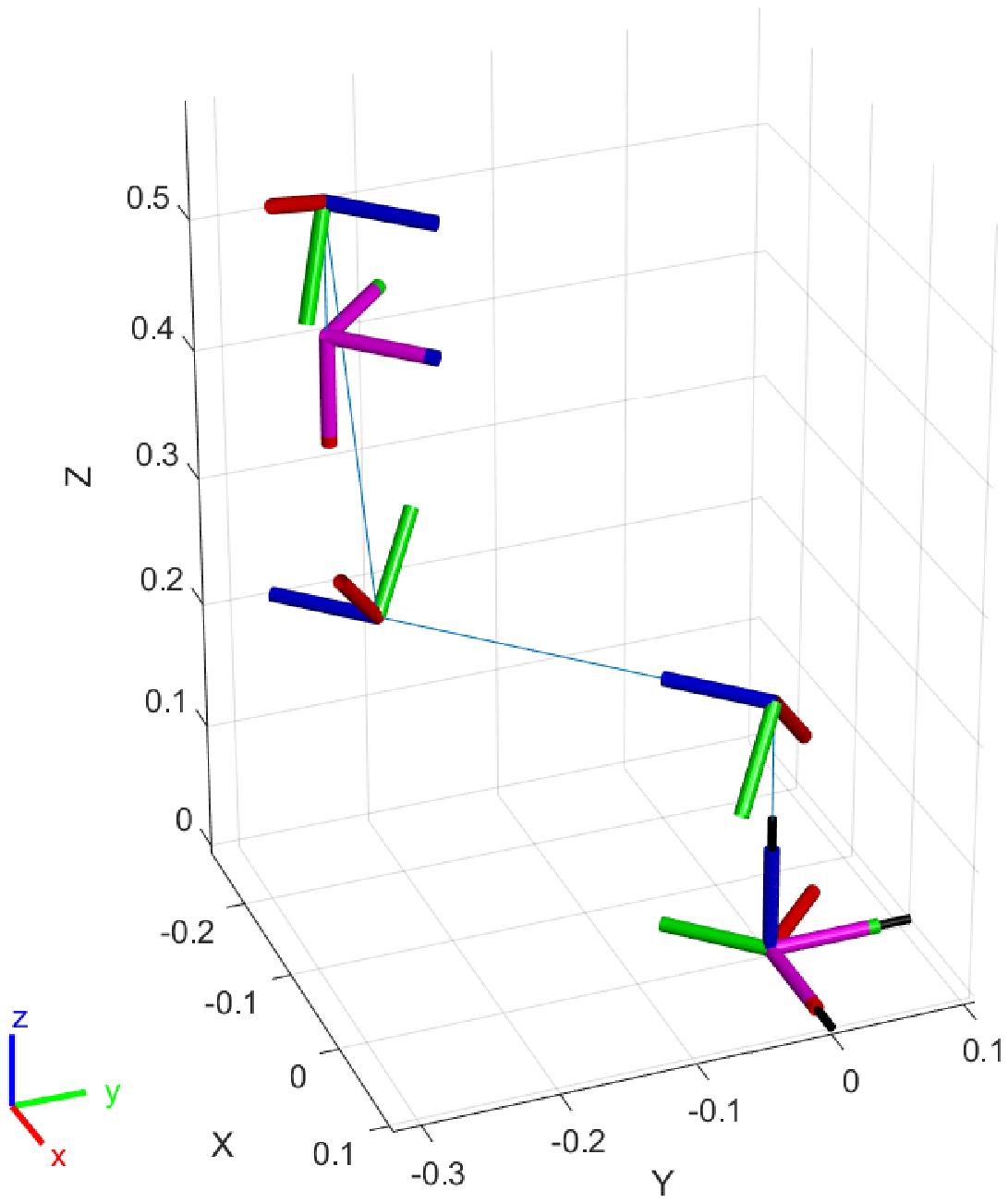}
    }
    \subfigure[Output DH parameters]
    {
        \includegraphics[width=0.31\textwidth]{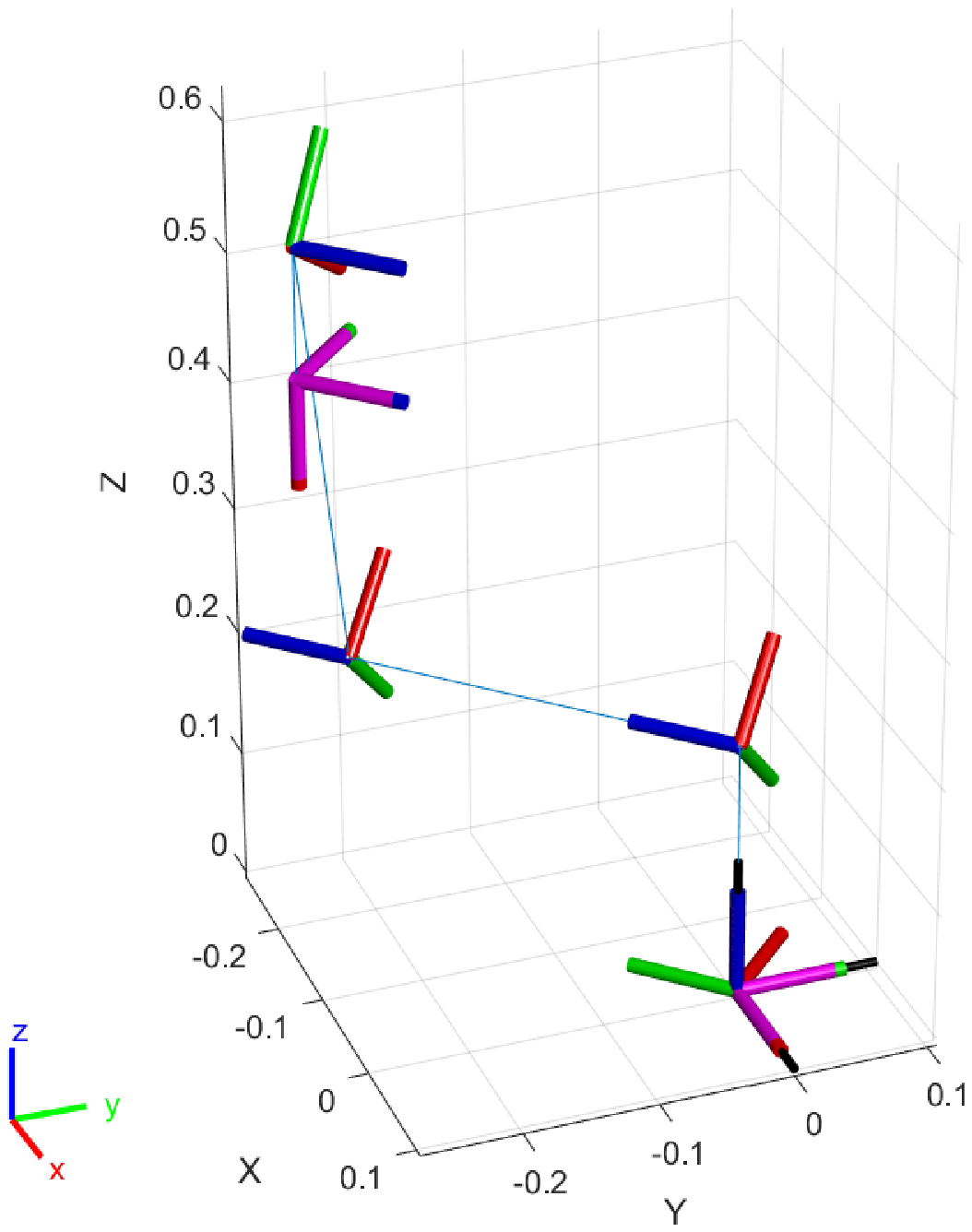}
    }
    \caption{The visualization of the RRPR robot in $[3\pi/4, -\pi/4, 0.3, -3\pi/4]^T$ configuration; joint axes have blue color; tool frame has pink color.}
     \label{fig:results_rrpr_p}
\end{figure*}

\begin{figure}
    \centering
    \includegraphics[width=0.48\textwidth]{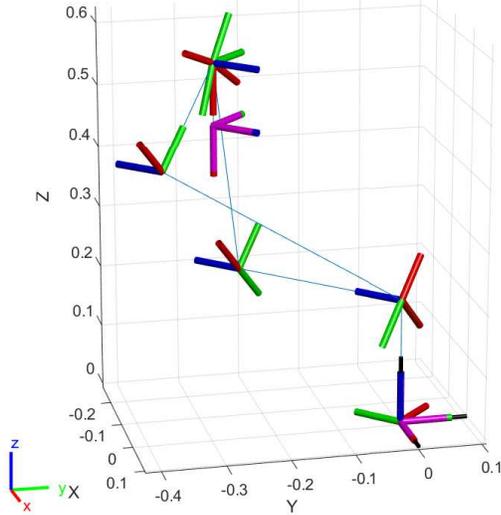}
    \caption{All 3 representations in one figure; RRPR robot configuration is $[3\pi/4, -\pi/4, 0.3, -3\pi/4]^T$; joint axes have blue color; tool frame has pink color. }
    \label{fig:results_rrpr_p-all}
\end{figure}

The prismatic joints in the PoE representation do not keep the moment of the screw axis, i.e., its position in space, because a prismatic displacement does not apply any moment on the origin. If one tries to obtain the DH parameters from those screws, the output parameters will differ in comparison with the original DH input. To see this limitation, we used the values from (\ref{eq:m_rrpr}) and (\ref{eq:screws_rrpr}) as input to generate another DH table of the same mechanism. The output is shown in Table \ref{tab:new_dh_RRPR} and is clearly different from Table \ref{tab:dh_RRPR}. However, the forward kinematics for these two DH representations yields the same result for the same input of joint variables. This is visualized in Figure \ref{fig:results_rrpr}, where the manipulator is in its home configuration, and in Figures \ref{fig:results_rrpr_p} and \ref{fig:results_rrpr_p-all} in $[3\pi/4, -\pi/4, 0.3, -3\pi/4]^T$ configuration.

\begin{table}[h]
\caption{Altered DH parameters of the RRPR robot (inputs $\mathbf{M}$ and $\mathcal{S}$).}
\vspace{-8pt}
\label{tab:new_dh_RRPR}
\begin{center}
\begin{tabular}{c | c c c c }
 i    &  $a_i$ [m] &  $d_i$ [m] &  $\alpha_i$ [rad] &  $\theta_i$ [rad] \\
 \hline\hline
 0 & 0 & 0 & 0 & 0 \\ 
 \hline
 1 & 0 & 0.2 & --$\pi$/4 & $q_1$(0) \\
 \hline
 2 & 0 & 0 & 0 & $q_2$(0) \\
 \hline
 3 & --0.361 & $q_3$(0) & $\pi$ & 2.159\\
 \hline
 4 & 0.1 & 0 & 0 & $q_4$(2.159)\\ 
\end{tabular}
\end{center}
\end{table}

\subsection{Arbitrary 3R structure}

This 3R robot is defined by PoE as follows.

\begin{equation}
\small
\label{eq:m_3r}
    \mathbf{M} = \\
    \begin{bmatrix}
        0.826 & -0.073 & -0.558 & 0.05 \\ 
        -0.373 & -0.814 & -0.444 & -0.4 \\
       -0.422 & 0.576 &  -0.699 & 0.4 \\
        0 & 0 & 0 & 1\\ 
    \end{bmatrix}
\end{equation}

\begin{equation}
\small
\label{eq:screws_3r}
    \mathcal{S}_1 = 
    \begin{bmatrix}
        -0.549 \\ 
        -0.099 \\
        0.829 \\ 
        0 \\
        0 \\ 
        0 \\
    \end{bmatrix}\quad
    \mathcal{S}_2 = 
    \begin{bmatrix}
        -0.635 \\ 
        0.495 \\
        0.592 \\ 
        -0.057 \\
        -0.182 \\ 
        0.090 \\
    \end{bmatrix}\quad
    \mathcal{S}_3 = 
    \begin{bmatrix}
        -0.280 \\ 
        0.790 \\
        0.544 \\ 
        -0.117 \\
        -0.206 \\ 
        0.238 \\
    \end{bmatrix}
\end{equation}

\begin{figure}
    \centering
    \includegraphics[width=0.48\textwidth]{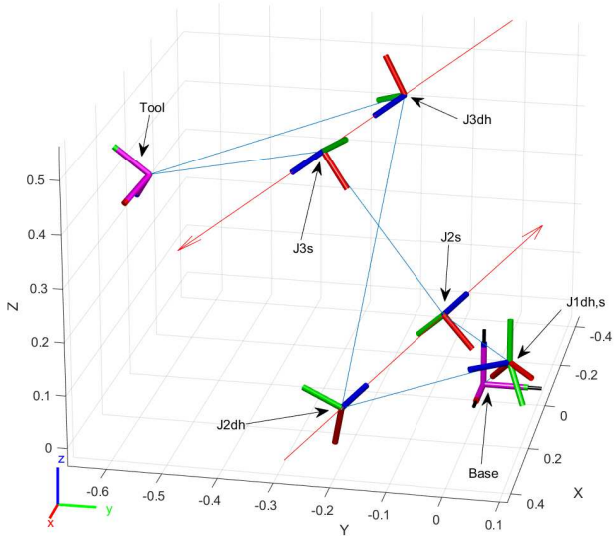}
    \caption{Visualization of the 3R robot representations at home configuration; the red arrows prove that the joint axes of DH and PoE representations are coincident, i.e. the kinematic structure is the same.}
    \label{fig:3r}
\end{figure}

The corresponding DH representation is shown in Table \ref{tab:dh_3r}, to provide the same forward kinematic results, a tool frame offset 
\begin{equation}
\small
\label{eq:T_3r}
    \mathbf{T}_t = \\
    \begin{bmatrix}
        0.651 &  -0.438 &  -0.619 & 0.105 \\
        0.653  &  0.739  &  0.163  &  0.394 \\
        0.386  & -0.511 &   0.767  & -0.121 \\
        0 & 0 & 0 & 1\\ 
    \end{bmatrix}
\end{equation}
has to be added to the final forward kinematics equation. The manipulator is visualized in Figure \ref{fig:3r} in its home configuration. The RPY-XYZ representation is shown in Table \ref{tab:rpy_3R}.

\begin{table}[h]
\caption{The DH parameters of the 3R robot}
\vspace{-8pt}
\label{tab:dh_3r}
\begin{center}
\begin{tabular}{c | c c c c }
 i    &  $a_i$ [m] &  $d_i$ [m] &  $\alpha_i$ [rad] &  $\theta_i$ [rad] \\
 \hline\hline
 0 & 0 &   0 &    --0.592 &    1.7502 \\ 
 \hline
 1 & --0.204&    0.088 &    0.658 &   $q_1$(1.758) \\
 \hline
 2 & --0.078  &   --0.325 &    0.467 &    $q_2$(--0.866) \\
 \hline
 3 & --0.515  &  0.314 &   --2.184 &    $q_3$(--1.743) \\ 
\end{tabular}
\end{center}
\end{table}

\begin{table}[h]
\caption{The RPY-XYZ of the 3R robot}
\vspace{-8pt}
\label{tab:rpy_3R}
\begin{center}
\begin{tabular}{c c c c c c}
 R [rad] &  P [rad] &  Y [rad] &  X [m] & Y [m] &  Z [m] \\
\hline\hline
0    &     0     &     0     &     0     &     0    &      0 \\ 
\hline
0.0998   & --0.5851     &     0   &       0     &     0     &     0 \\ 
\hline
0.6423 &    0.1577  &  --3.0111  &   0.2071  &  0.0272    & 0.0332 \\ 
\hline
0.3616 &   --0.3037  &   0.0622  &  --0.1089   & --0.0199  &  --0.1006 \\ 
\hline
--2.6489  &   0.8582  &  --2.5611  &   0.1168  &   0.5115 &   --0.1124 \\
\end{tabular}
\end{center}
\end{table}

\section{Conclusion}

This paper presents a set of analytical algorithms which can interpret between three commonly used kinematic representations -- the Standard Denavit-Hartenberg notation, Roll-Pitch-Yaw angles with XYZ displacement, and the Screw Theory approach (Product of Exponentials). It contributes to the community of roboticists who have to deal with inconvenient switching between those representations for multiple reasons, such as software compatibility, methodology simplification, or team cooperation. Under some stated limitations, such as placement of prismatic joints in case of PoE, the method can convert the representations back and forth multiple times with the possibility of generating a simple version of a URDF file. The most common joints, revolute and prismatic, are supported. Due to the analytical approach, the singularities do not cause a problem during conversion -- the only key aspect is the correct formulation of a robot in its home configuration via an input representation. MATLAB scripts and implemented examples are available as an open source package in \cite{Huczala2021git}. The main advantage of this novel approach is to represent the joint coordinates of a robot globally as a matrix form of the \textit{SE}(3) group with respect to the base of the robot. From this interpretation, it is possible to derive any other commonly used kinematic representation when all of them operate as rigid body motions of \textit{SE}(3). The future development involves the creation of a Python version with GUI, an extension to the Modified DH representation, and the change of the screw coordinate frame using adjoint matrices.

\section*{Acknowledgment}

The cooperation between the authors was initiated thanks to the \textit{Aktion \"{O}sterreich-Tschechien Semesterstipendien} grant, financed by Federal Ministry of Education, Science and Research (BMBWF) and organized by Austrian Agency for International Cooperation in Education and Research (OeAD).

\bibliographystyle{IEEEtran}
\bibliography{references}

\end{document}